\documentclass[11pt]{article}

\usepackage[margin=1in]{geometry}
\usepackage{graphicx}
\usepackage{multirow}
\usepackage{amsmath,amssymb,amsfonts}
\usepackage{booktabs}
\usepackage{textcomp}
\usepackage{xcolor}
\usepackage{url}
\usepackage[LGR,T1]{fontenc}
\usepackage[utf8]{inputenc}
\usepackage{textalpha}
\usepackage[round,authoryear]{natbib}
\usepackage{authblk}
\usepackage{hyperref}

\title{Svarna: An Open Corpus Workbench for Modern Greek}
\author{Stergios Chatzikyriakidis}
\affil{Department of Philology, University of Crete, Rethymno, Greece}
\affil{\texttt{stergios.chatzikyriakidis@gu.se}}
\date{}

\begin{document}

\maketitle

\begin{abstract}
This paper introduces Svarna, a free, open-source, web-based corpus workbench for Modern Greek. Svarna integrates five databases covering various registers, institutional, literary, dialectal, social media, and historical, to provide a total of more than 507 million words and around 29 million sentences. This platform attempts to address a chronic problem in Greek language technology. Although various corpus resources exist, they are scattered across different platforms, and in many cases, institutional access is restricted or they are no longer available online. Svarna integrates a number of these resources into a single interface that can be used without logging in, installation, or specialized training. SVARNA  provides a concordancer with KWIC marking capabilities, frequency analysis including register-by-register normalization, collocation extraction using mutual information, a dictionary of 93 Greek discourse markers providing distribution profiles, text-level analysis tools including n-grams, varieties, and collocation networks, register comparison using log-ratio, regular expression search, and an optional LLM layer for pragmatic annotation and free research mode. This platform is built upon SQLite FTS5 full-text indices provided via a FastAPI backend, deployed as Docker containers on Azure, and released under the MIT license. Source code, build scripts, and deployment configurations are publicly available on GitHub. What is important, is that users can easily ingest their own corpora and deploy their own instances. This paper describes the system design, corpus structure, and use cases demonstrating the various queries supported by the platform. Svarna serves as the first step in exploring available data and aspires to lay the foundation for more comprehensive research in the future.
\end{abstract}


\section{Introduction}\label{sec:intro}
	Modern Greek is spoken by approximately 13 million people and possesses a written tradition that has been continuously used for centuries. However, the Greek corpus infrastructure is surprisingly fragmented. Researchers, students, and language experts who wish to search, query, and analyze Greek texts on a large scale face difficulties due to inaccessible and scattered materials.
	
	Of course, materials are not entirely nonexistent. Over the past 20 years, valuable resources on the Greek language have been produced through various corpus construction projects. The Hellenic National Corpus (HNC) by ILSP/Athena RC was one of the early large-scale projects \citep{hatzigeorgiu2000, mikros2005}. The Corpus of Greek Texts (CGT) built by \cite{goutsos2010} provides a balanced reference corpus. The CLARIN:EL infrastructure \citep{gavriilidou2023} provides various language resources through national portals. Greek data is included in large-scale multilingual datasets such as CC-100 \citep{conneau2020}, OSCAR \citep{ortizsuarez2019}, and mC4 \citep{xue2021}. Similar materials are also available in Europarl \citep{koehn2005} and OpenSubtitles \citep{lison2016}. The Leipzig Corpora Collection contains Greek web data \citep{goldhahn2012}, and Greek treebanks exist in Universal Dependencies \citep{prokopidis2017, nivre2016}. Parliamentary minutes have been constructed as a dataset \citep{dritsa2022}, and recently, dialect data has been collected from GRDD and GRDD+ datasets \citep{chatzikyriakidis2023grdd, chatzikyriakidis2026grddplus}.
	
	The problem is not a lack of resources, but rather the difficulty in utilizing existing resources together. Access to some corpora is restricted to institutional credentials that are not available to everyone. Some platforms that once offered search capabilities are no longer reliably available online. Other platforms require the installation of specialized software or knowledge of specific languages, which most users lack. There is no single platform that allows Greek scholars, language teachers, translators, or students to easily search Greek texts spanning various tones, genres, and dialects.
	
	This paper presents Svarna\footnote{The name comes from the Greek σβάρνα, a traditional agricultural tool used for leveling and smoothing the ground after plowing. We use it metaphorically: this work is a first sweep over the available resources, intended to level the ground for further, more refined efforts.} (Σβάρνα), an open corpus workbench for Modern Greek that attempts to address this gap. Svarna is a web-based platform that integrates five independently searchable databases covering institutional language (Wikipedia, parliamentary proceedings, web crawl data, subtitles, parallel corpora, and treebank data; roughly 473 million words), literary and developmental language (prose from Project Gutenberg, child-directed input from BabyLM, interwar poetry, and the Tesserae classical Greek corpus; 26.6 million words), dialectal text (nine Greek dialect groups from the GRDD+ collection; 5.9 million words), social media (Greek Twitter/X data labelled for toxicity; 1.9 million words), and historical/folkloric text (CLARIN:EL Crete collections). The total is more than 507 million words across roughly 29 million sentences.
	
	This platform is completely free and open. It requires no login, API keys (except for optional LLM features), and no installation. This system is distributed as a web application accessible to anyone. The source code, database build scripts, and deployment configurations are all publicly available on GitHub under the MIT license. It is designed to allow anyone to add new corpora, extend functionality, or deploy their own instances.
	
	Svarna is not intended to replace existing corpus analysis tools for resource-rich languages, such as Sketch Engine \citep{kilgarriff2004, kilgarriff2014}, CQPweb \citep{hardie2012}, or AntConc \citep{anthony2005}. Nor does it aim to compete with the meticulous management of purpose-built reference corpora. Rather, it aims to consolidate already available free resources into a single interface for searching and provide the analysis capabilities expected by corpus linguists. It would be even better if more comprehensive and well-organized resources on Greek were to emerge in the future. In the meantime, Svarna provides a useful tool.

	\section{Related Work}\label{sec:related}
	
	\subsection{Greek Corpus Resources}\label{subsec:greek-corpora}
	
	The Hellenic National Corpus (HNC), developed by the ILSP/Athena Research Centre, was one of the first large-scale Greek corpora containing approximately 47 million words extracted from various written forms \citep{hatzigeorgiu2000}. Quantitative research on Greek based on the HNC established fundamental distributional characteristics of the language, including word length distributions and frequency characteristics \citep{mikros2005, hatzigeorgiu2001}.
	
	The Corpus of Greek Texts (CGT) was designed as a balanced reference corpus representing various forms of written language \citep{goutsos2010}, and its impact on research is significant. However, online access to the CGT is unstable, and search interfaces are not always available to external users.
	
	At the national level, the CLARIN:EL infrastructure \citep{gavriilidou2023} provides a portal of Greek resources including corpora, dictionaries, and natural language processing tools. While CLARIN:EL represents a significant infrastructure investment, an institutional account is required to access much of the material, and navigating the platform can be difficult for users who simply want to perform text searches.
	In addition to projects dedicated to the Greek language, Greek text is included in several large-scale multilingual collections. CC-100 \citep{conneau2020} contains Greek web data filtered from Common Crawl, and OSCAR \citep{ortizsuarez2019} provides classified and filtered web text for Greek. The mC4 dataset \citep{xue2021} also includes Greek. However, these materials are primarily designed for training language models and are not suitable for language search and analysis. Generally, these materials are distributed in the form of raw text without indexing or analysis tools.
	
	Parallel corpora containing Greek are available through Europarl \citep{koehn2005}, which sorts the minutes of the European Parliament by language, and OPUS \citep{tiedemann2012}, which integrates parallel data from various sources including OpenSubtitles \citep{lison2016}. The Leipzig Corpora Collection \citep{goldhahn2012} provides Greek data collected via web crawling that includes sentence-level frequency information. The Greek treebank of the Universal Dependencies framework \citep{nivre2016} provides data with morphological and syntactic annotations, and this also includes the Greek Dependency Treebank \citep{prokopidis2017}.
	
	Recently, parliamentary minutes have been constructed into structured datasets suitable for computational analysis \citep{dritsa2022}. Greek data was also used in the BabyLM collaboration project \citep{warstadt2023}, which provides child-centered and child-accessible texts for language modeling experiments. Greek dialect data was collected and released through the GRDD and GRDD+ projects \citep{chatzikyriakidis2023grdd, chatzikyriakidis2026grddplus} and covers nine dialect groups, including Cretan, Cypriot, Pontic, Tsakonian, and Griko.
	
	A survey of Greek natural language processing (NLP) resources \citep{papantoniou2020, papantoniou2024} identified the current state of the field and pointed out persistent gaps, particularly the lack of searchable and publicly available multi-register corpora. Recent evaluation studies have further emphasized the need for comprehensive test resources \citep{kogkalidis2024}, and Greek language models such as GreekBERT \citep{koutsikakis2020} have shown potential to leverage available text data.
	
	\subsection{Corpus Linguistics Tools}\label{subsec:tools}
	
	Tools primarily used in corpus linguistics offer various combinations of functions such as word search, frequency analysis, collocation extraction, and distribution analysis. Sketch Engine \citep{kilgarriff2004, kilgarriff2014} is a commercial platform that provides word sketching, thesaurus features, and corpus building tools, supporting various languages. CQPweb \citep{hardie2012} provides a web-based interface to IMS Corpus Workbench (CWB), combining the powerful capabilities of CQP query syntax with browser accessibility. Voyant Tools \citep{rockwell2016} is a web-based text analysis environment specialized for digital humanities, offering visualization features such as word clouds, trend analysis, and document-level statistics. AntConc \citep{anthony2005} is a desktop application for corpus search that is widely used, particularly in the fields of language education and applied linguistics.
	
	While each of these tools has its advantages, there is no tool that comes pre-loaded with Modern Greek data to enable immediate searching across multiple dialects and diverse vocabularies without separate configuration. For researchers or students interested in Greek, the process from the idea to actually obtaining results typically requires downloading data, installing software, building an index, and learning search phrases. Svarna aims to eliminate these inconveniences.

	\section{System Architecture}\label{sec:architecture}
	
\begin{figure}[htbp]
		\centering
		\includegraphics[width=0.85\textwidth]{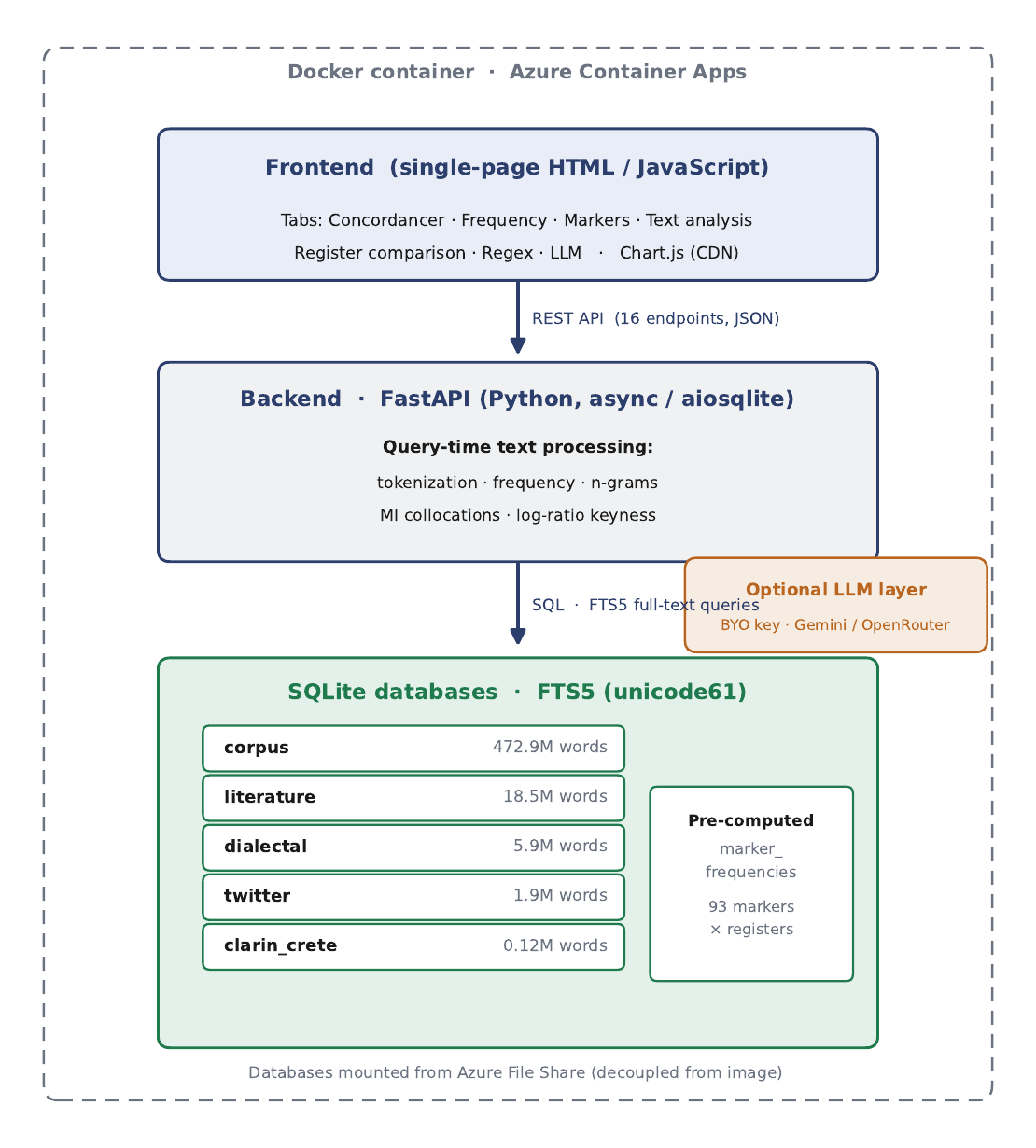}
		\caption{Svarna system architecture. A single-page frontend communicates with a FastAPI backend over a REST API; the backend serves the SQLite FTS5 databases and performs text processing at query time. An optional, bring-your-own-key LLM layer sits alongside the backend. Code and data are decoupled: the databases are mounted from an Azure File Share into the Docker container.}
		\label{fig:architecture}
	\end{figure}

	Svarna follows a simple client-server architecture (Figure~\ref{fig:architecture}). The backend is a Python application built on FastAPI that serves a single-page HTML/JavaScript frontend. All corpus data is stored in SQLite databases using the FTS5 (Full-Text Search 5) extension, which provides efficient full-text indexing with the \texttt{unicode61} tokenizer for proper handling of Greek Unicode characters.
	
	Each corpus is a self-contained SQLite FTS5 database, and the build scripts are released with the source code, which makes the system straightforward to extend; the procedure for adding a new corpus is described in Section~\ref{subsec:extending}.
	
	\subsection{Database Design}\label{subsec:db}
	
	Each database consists of a single SQLite file containing an FTS5 virtual table called \texttt{sentences}. The rows of this table represent sentences and include JSON-formatted metadata fields containing the sentence body, register labels (e.g., Wikipedia, Parliament, subtitle, poem, Cretan), and source-specific information (e.g., title, author, dialect group, etc.).
	
	Using FTS5 indexes, efficient phrase searching, prefix searching, and Boolean queries can be performed directly via SQL. FTS5 operates at the token level and natively supports Unicode normalization, so there are no issues with parsing Greek letters and diacritics.
	
	In addition to FTS5 indexes, each database includes a pre-calculated \texttt{marker\_frequencies} table that stores the frequencies of 93 Greek discourse markers classified by register.
	
	The system supports five databases: the Corpus Database, the Literature Database, the Dialectal Database, the Social Media Database, and the Historical Database.
	
	\subsection{Backend}\label{subsec:backend}
	
	The backend provides sixteen REST API endpoints supporting word match search, frequency analysis, collocation extraction, discourse marker search, text analysis, regular expression search, register list creation, database enumeration, and more. All database access is performed asynchronously, and aiosqlite is used to prevent event loop blocking during queries.
	
	The key here is that all text processing (e.g.\ tokenization, frequency calculation, n-gram extraction, and collocation score calculation) is performed in Python at query time instead of relying on pre-calculated indexes. This keeps the database simple and speeds up index building, but response times may be somewhat slower for computationally intensive queries, such as collocation extraction for very high-frequency words.
	
	Collocation scores are calculated using pointwise mutual information \citep{church1990}.
	
	\begin{equation}
		\text{MI}(x, y) = \log_2 \frac{p(x, y)}{p(x) \cdot p(y)}
	\end{equation}
	
	\noindent Here, $p(x, y)$ is estimated from the co-occurrence frequency within the window, and $p(x)$ and $p(y)$ are estimated from the marginal frequencies of the sampled match rows.
	
	For register comparison, the system calculates the log-ratio score \citep{rayson2000} to identify words that appear statistically over-represented in one register compared to another, according to the keyness analysis methodology described by \cite{scott1997}.
	
	\subsection{Frontend}\label{subsec:frontend}
	
	The frontend is essentially a single HTML file (approximately 3,500 lines) containing built-in JavaScript and CSS. It does not use build tools, frameworks, or package managers. Chart.js is loaded from a CDN for visualization. The interface consists of tabs corresponding to the core analysis functions: match analysis, frequency analysis, discourse markers, text analysis, regular expression search, register comparison, and LLM analysis.
	
	The design philosophy is intentional minimalism. All functions run in the browser. It uses pure JavaScript. Therefore, the application can be easily modified and deployed, and users who wish to extend it can also easily understand and utilize it.
	
	\subsection{Deployment}\label{subsec:deployment}
	
	This application is containerized into a Docker container and deployed to Azure Container Apps. The five database files are stored in an Azure file share mounted to the container. They are mounted at runtime. Since the code and data are separated, the database can be updated without rebuilding the container image. The deployment script, Dockerfile, and Azure configuration files are included in the repository.
	
	The actual instance is publicly available to users for free.\footnote{Available at \url{https://greek-corpus-workbench.wonderfulhill-e1c9f1a0.westeurope.azurecontainerapps.io/}}
	
	\subsection{Adding New Corpora}\label{subsec:extending}
	
	Each corpus is a self-contained SQLite FTS5 database, and since the build script is provided with the source code, you do not need to change the application to extend Svarna. Adding a corpus consists of the following three steps: (i) get source text; (ii) run a build script that splits the text into sentences and stores them in a database using a standard schema (\texttt{sentences} FTS5 table, \texttt{corpus\_stats}, and a pre-calculated \texttt{marker\_frequencies} table); then (iii) register the database via environment variables. The registered database appears in the navigation bar selector alongside the existing corpus.
	
	To illustrate this, we provide a support pipeline (\texttt{svarna-glossapi}) for the glossAPI initiative \citep{glossapi}, an EELLAK/GFOSS project. glossAPI is a project that releases Greek datasets from various fields, including academia, law, journalism, and literature, under a Creative Commons license. This pipeline takes a manifest as input that maps each dataset to a topic, style, and mode. For each dataset, it streams data from HuggingFace Hub, automatically detects text fields, splits them into sentences using the same rules as the core builder, and then creates a Svarna database for each topic. Datasets containing only metadata (e.g., link indexes without body text) are automatically detected and skipped. As a proof of concept, we built a literature database consisting of approximately 556,000 sentences and 11 million words from six glossAPI collections (Drama, Classics, Early Greek, Ecclesiastical Literature, Wikisource, and Folklore). Since this procedure applies to all glossAPI datasets, the scope of Svarna can be extended to academic, legal, and media fields without additional engineering work. The pipeline and manifest are distributed under the MIT license.

	\section{Corpora}\label{sec:corpora}
	
	Svarna integrates data from multiple freely available sources into five thematic databases. Table~\ref{tab:corpora} summarizes the composition.
	
	\begin{table}[htbp]
		\caption{Overview of the five Svarna databases.}\label{tab:corpora}
		\begin{tabular*}{\textwidth}{@{\extracolsep\fill}llrrl}
			\toprule
			Database & Register / Source & Sentences & Words & Source \\
			\midrule
			\multirow{16}{*}{Corpus}
			& CC-100 (web)            & 5,000,000  & 112,135,259 & \cite{conneau2020} \\
			& Wikipedia (encyc.)      & 4,739,089  & 67,569,423  & Wikimedia dumps \\
			& Europarl (instit.)      & 1,589,886  & 39,342,006  & \cite{koehn2005} \\
			& OPUS WikiMatrix         & 2,000,000  & 39,317,133  & \cite{tiedemann2012} \\
			& OPUS EUbookshop         & 2,000,000  & 36,233,670  & \cite{tiedemann2012} \\
			& OPUS ParaCrawl          & 2,000,000  & 31,004,648  & \cite{tiedemann2012} \\
			& Parliament (political)  & 198,850    & 24,561,364  & \cite{dritsa2022} \\
			& Leipzig news 2020       & 1,000,000  & 19,075,841  & \cite{goldhahn2012} \\
			& Leipzig newscrawl 2011  & 999,996    & 19,743,123  & \cite{goldhahn2012} \\
			& Leipzig newscrawl 2017  & 1,000,000  & 19,452,653  & \cite{goldhahn2012} \\
			& Leipzig wikipedia 2016  & 1,000,000  & 18,903,388  & \cite{goldhahn2012} \\
			& Leipzig wikipedia 2021  & 1,000,000  & 18,060,085  & \cite{goldhahn2012} \\
			& OPUS CCAligned (web)    & 2,000,000  & 17,370,501  & \cite{tiedemann2012} \\
			& OpenSubtitles (dialogue)& 2,000,000  & 10,053,036  & \cite{lison2016} \\
			& UD Greek-GDT            & 2,521      & 55,633      & \cite{prokopidis2017} \\
			& UD Greek-GUD            & 1,807      & 20,536      & \cite{nivre2016} \\
			\midrule
			\multirow{4}{*}{Literature}
			& Gutenberg & 562,432 & 8,227,594 & Project Gutenberg \\
			& BabyLM & 1,049,266 & 10,177,729 & \cite{warstadt2023} \\
			& Poetry & 4,087 & 110,823 & Interwar poetry \\
			& Tesserae (Anc. Greek) & 310,928 & 8,131,811 & Tesserae Project \\
			\midrule
			\multirow{9}{*}{Dialectal}
			& Cretan                  & 150,000    & 1,600,000   & \multirow{9}{*}{\cite{chatzikyriakidis2026grddplus}} \\
			& Cypriot                 & 120,000    & 1,300,000   & \\
			& Pontic                  & 80,000     & 900,000     & \\
			& Northern                & 60,000     & 650,000     & \\
			& Eptanisian              & 40,000     & 450,000     & \\
			& Tsakonian               & 35,000     & 400,000     & \\
			& Maniot                  & 25,000     & 250,000     & \\
			& Griko                   & 20,000     & 200,000     & \\
			& Katharevousa            & 12,935     & 148,147     & \\
			\midrule
			\multirow{2}{*}{Social media}
			& Twitter (non-toxic)     & 58,294     & 1,222,768   & \multirow{2}{*}{Twitter / X} \\
			& Twitter (toxic)         & 30,389     & 674,936     & \\
			\midrule
			\multirow{2}{*}{Historical}
			& CLARIN Crete (13 colls.)& 3,599      & 117,004     & CLARIN:EL \\
			&                         &            &             & \cite{gavriilidou2023} \\
			\midrule
			\multicolumn{2}{l}{\textbf{Total}} & \textbf{29,094,079} & \textbf{507,459,111} & \\
			\bottomrule
		\end{tabular*}
	\end{table}
	\subsection{Corpus Database}\label{subsec:corpus-db}
	
	The main corpus database includes institutional and general-purpose Greek. The largest component is the Greek Wikipedia, an encyclopedic prose covering a wide range of topics. The Greek Parliament Minutes \citep{dritsa2022} provides official political discourse with identified speakers. Web crawled data from CC-100 adds diversity in terms of registers and topics. OpenSubtitles provides informal colloquial expressions extracted from movie and TV dialogue. Europarl provides official EU discourse. The Leipzig Corpus Collection adds sentences collected from the web. The Greek treebank data from Universal Dependencies provides texts that are smaller in volume but morphosyntactically rich.
	
	Each sentence displays the source register, allowing users to filter searches and compare distributions across various registers.
	
	\subsection{Literature Database}\label{subsec:lit-db}
	
	This literature database focuses on creative and humanistic Greek texts. The largest part is 221 Greek books provided by Project Gutenberg (novels, essays, and historical works). The BabyLM component \citep{warstadt2023} includes Greek texts extracted from a children's corpus and uses concise and easy-to-understand language. Interwar Greek Poetry (1920s--1940s) adds poetry from a historically significant literary period. It also incorporates the Tesserae classical Greek corpus (820 works; roughly 310,000 sentences and 8.1 million words), enabling diachronic comparison between ancient and modern Greek.
	
	The metadata for each sentence includes information on the title, author, or collection, allowing users to verify the source of individual search results.
	
	\subsection{Dialectal Database}\label{subsec:dial-db}
	
	The dialectal database contains data from the GRDD+ collection \citep{chatzikyriakidis2023grdd, chatzikyriakidis2026grddplus}, which is the most comprehensive publicly available dataset of Greek dialectal text. It covers nine dialect groups: Cretan, Cypriot, Pontic, Northern Greek, Eptanisian (Ionian Islands), Tsakonian, Maniot, Griko (Italiot Greek), and Katharevousa (the formal written register used until the mid-20th century, included for historical comparison).
	
	This database enables comparative queries across dialects, such as searching for the same lexical item or construction and observing its distribution by dialect group.
	
	\section{Features}\label{sec:features}
	
	Svarna provides various analytics functions organized into tabs in a web interface. Each function can be accessed with a single mouse click, and no separate search language is required.
	
	\subsection{Concordancer}\label{subsec:kwic}
	
	The Concordancer performs KWIC (Key Word in Context) searches and allows you to set the context width. In search results, the search term is highlighted in the center column, with contextual information displayed on the left and right.
	
	Searching uses FTS5 phrase matching, which supports multi-word queries, prefix queries (e.g., γλωσσ* matches all words starting with the stem), and Boolean operators.
	
	\begin{figure}[htbp]
		
		\centering
		
		\includegraphics[width=\textwidth]{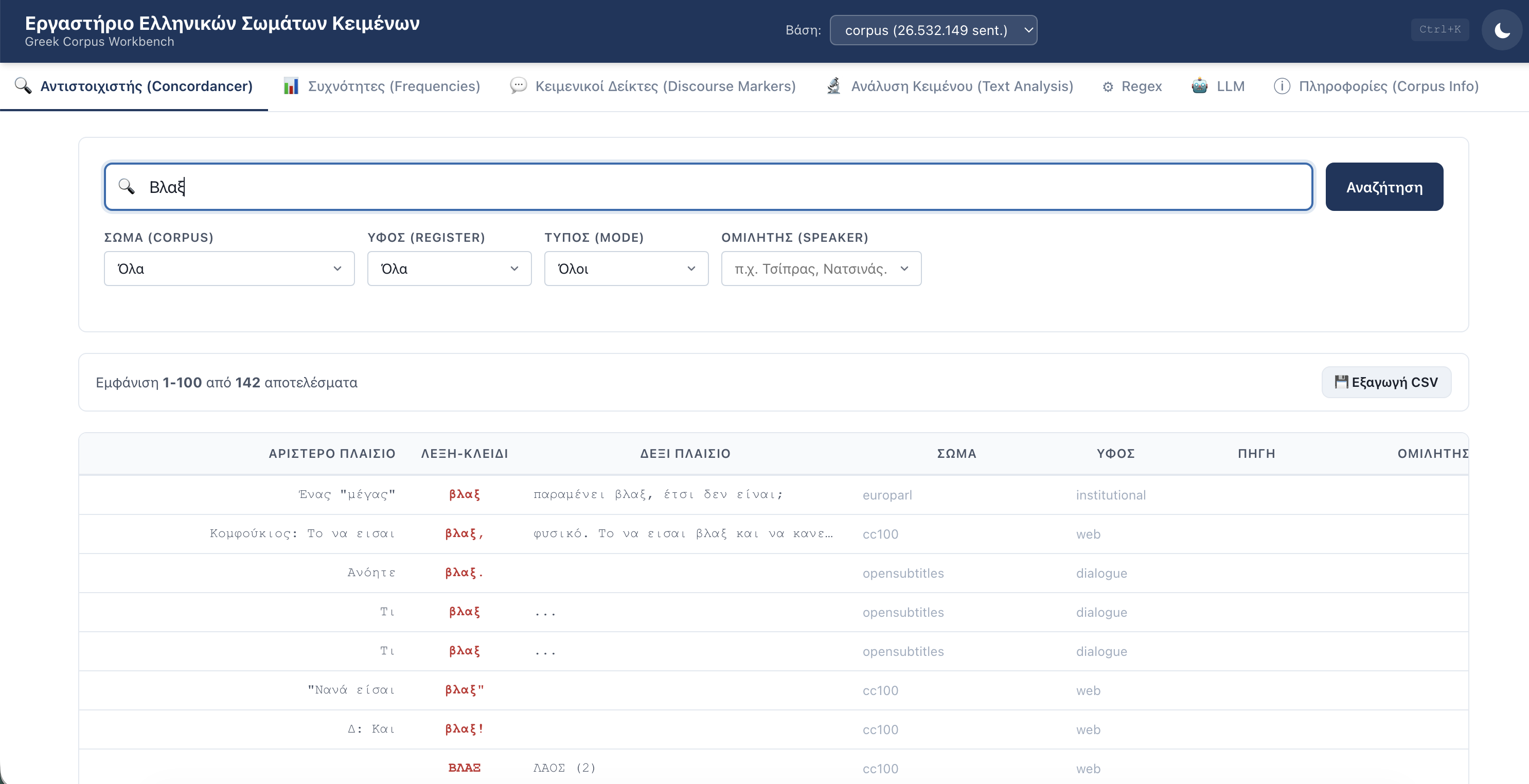}
		
		\caption{KWIC (Concordancer) interface. The search term is highlighted in the center column, along with contextual information on the left and right. Each row is labeled with Corpus, Register, Source, and Speaker. Results can be filtered by Corpus, Register, and Mode, and can be exported as a CSV file.}
		
		\label{fig:kwic}
		
	\end{figure}
	
	\subsection{Frequency Analysis}\label{subsec:freq}
	
	The Frequency tab provides the raw frequency and normalized frequency per million for the search term across all registers in the active database. Results are displayed in tables and bar graphs. Normalization per million enables meaningful comparisons between registers using token totals per register.
	
	Collocations are extracted from match search results using the Mutual Information (MI) scoring method \citep{church1990, evert2008}. For each collocation, the system reports co-occurrence frequency, MI score, and the individual frequencies of the node word and the collocate. The results are presented in the form of a rank table and a force-directed collocation network visualization.
	
	\begin{figure}[htbp]
		
		\centering
		
		\includegraphics[width=\textwidth]{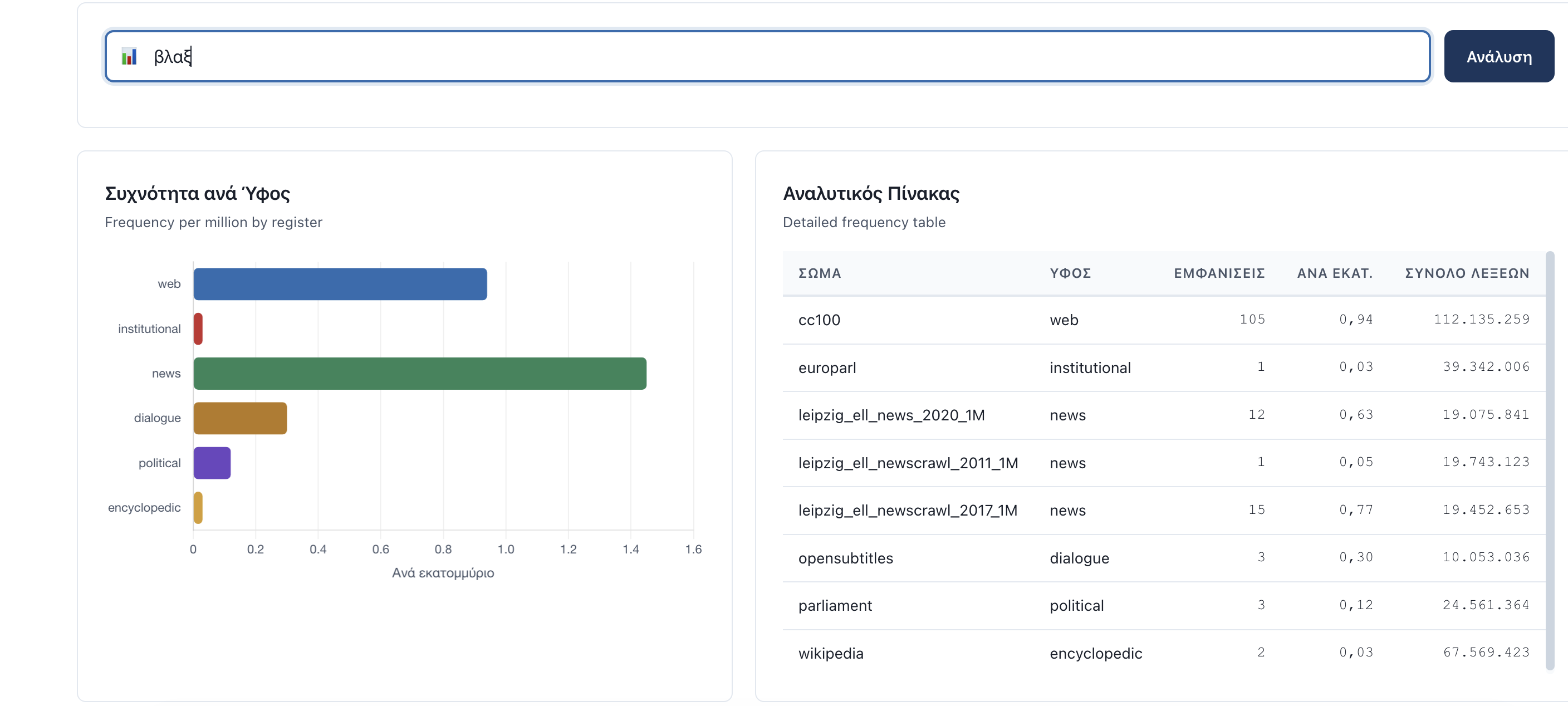}
		
		\caption{Frequency tab. The million-unit normalized frequency of the search term is displayed as a bar graph by register (left), along with a detailed table (right) providing raw frequency, million-unit ratio, and register token totals.}
		
		\label{fig:frequency}
		
	\end{figure}
	\subsection{Discourse Markers}\label{subsec:markers}
	
	The discourse marker tab provides access to a pre-computed inventory of 93 Greek discourse markers organized into 11 functional categories: causal (επειδή, γιατί, αφού, διότι), sequential (μετά, έπειτα, στη συνέχεια), additive (επίσης, ακόμα, επιπλέον), adversative (αλλά, όμως, ωστόσο), reformulative (δηλαδή, με άλλα λόγια), conclusive (λοιπόν, επομένως, άρα), topic management (τέλος πάντων, όσο αφορά), hedging (βέβαια, ίσως), interactional (ε, ρε, κοίτα, ξέρεις), evidential (φαίνεται, λένε, δήθεν), and conditional (αν, εάν).
	
	The classification draws on the discourse marker literature for Greek \citep{schiffrin1987, georgakopoulou1998} and provides distributional data that would otherwise require extensive manual corpus work. For each marker, the system shows its frequency by register, a bar chart of the distribution, and the option to jump to concordance lines for that marker.
	
	\begin{figure}[htbp]
		\centering
		\includegraphics[width=\textwidth]{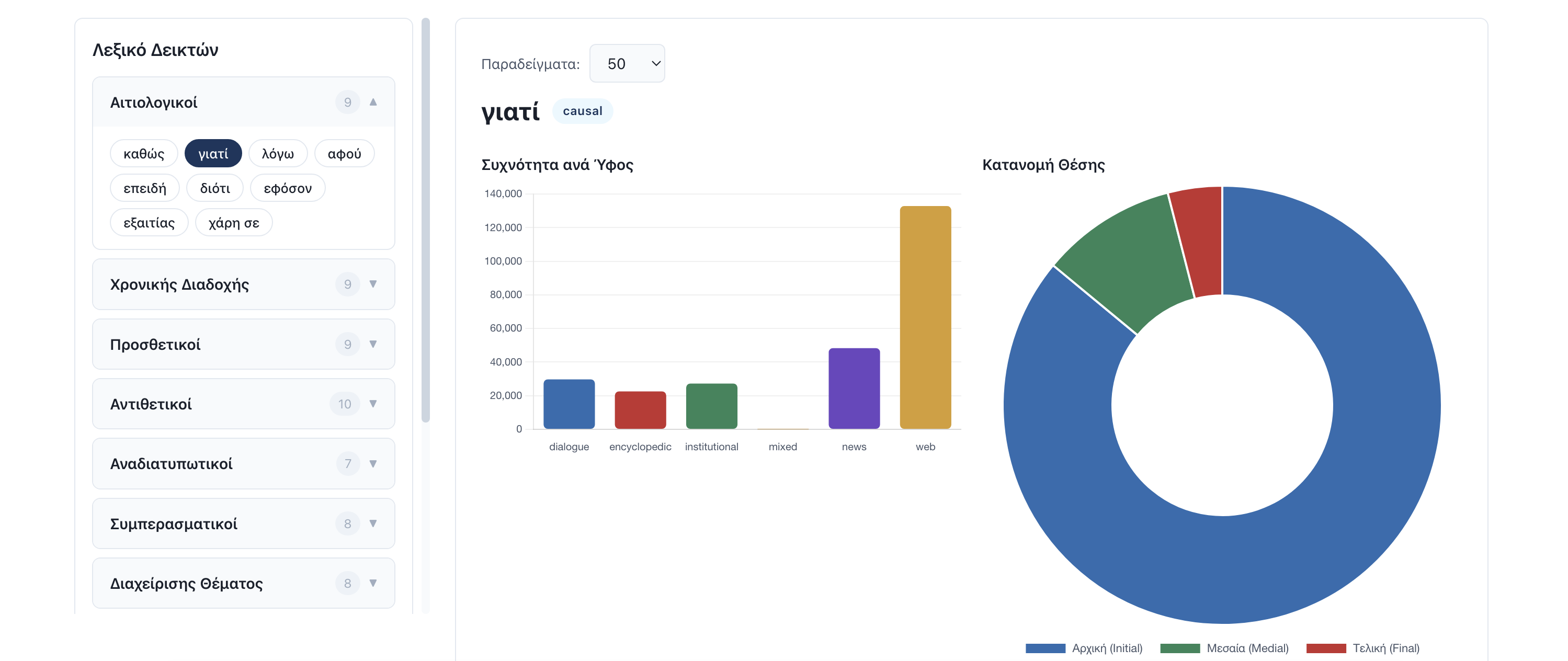}
		\caption{The discourse-markers tab. The lexicon of 93 markers is browsable by functional category (left); for a selected marker the system shows its frequency by register (centre) and a positional-distribution breakdown into initial, medial, and final position (right).}
		\label{fig:markers_ui}
	\end{figure}
	
	\subsection{Text Analysis, Register Comparison and Regex Search}\label{subsec:textanalysis}
	
	The text analysis tab provides Voyant-style functionality \citep{rockwell2016} over the corpus data. For a given search term, it generates word frequency lists ranked by raw count or TF-IDF, bigram and trigram frequency lists, a dispersion plot showing the distribution of the term across the concordance sample, and a force-directed collocation network.
	
	\begin{figure}[htbp]
		\centering
		\includegraphics[width=\textwidth]{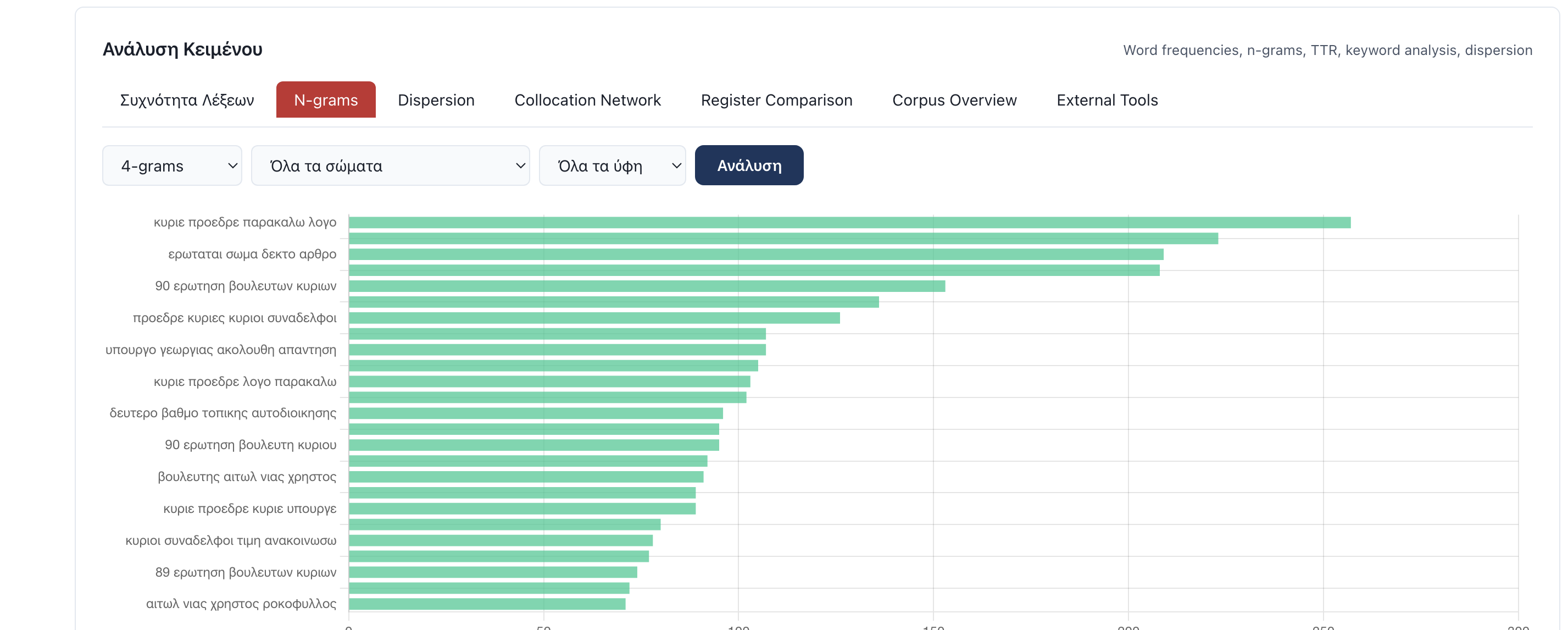}
		\caption{The text-analysis tab, here showing the most frequent 4-grams in the corpus. The sub-tabs (word frequencies, n-grams, dispersion, collocation network, register comparison, corpus overview) provide Voyant-style exploration over the indexed data.}
		\label{fig:ngrams}
	\end{figure}
	
	The register comparison feature allows users to select two registers and identify the words that are most characteristic of each, using log-ratio keyness scores \citep{rayson2000, scott1997}. This is useful for contrastive register studies, e.g., comparing parliamentary with subtitle language, or Wikipedia with poetry.
	
	A regex search tab provides raw regular expression matching across the corpus. This is intended for users with more technical requirements, such as searching for morphological patterns, specific character sequences, or orthographic variants.
	\subsection{LLM Integration}\label{subsec:llm}
	
	This app allows you to submit search results or text excerpts to an LLM. This enables pragmatic classification, discourse analysis, or corpus-based question answering. Users must provide their own API keys for this feature, and it is not included in the core corpus features.
	
	In addition to fixed analysis capabilities, Svarna provides optional LLM layers to allow users to explore natural language corpora. Users can provide their own API keys and ask questions using the LLM of their choice. In free research mode, users can ask open-ended questions about search results. For example, users can ask the model to summarize how specific markers are used at various registers, or to propose hypotheses about collocation patterns. These questions are based on actual retrieved data, not the model's parameter memory. Since the keys are user-owned, this feature does not incur additional costs or platform access restrictions.
	
	\begin{figure}[htbp]
		
		\centering
		
		\includegraphics[width=\textwidth]{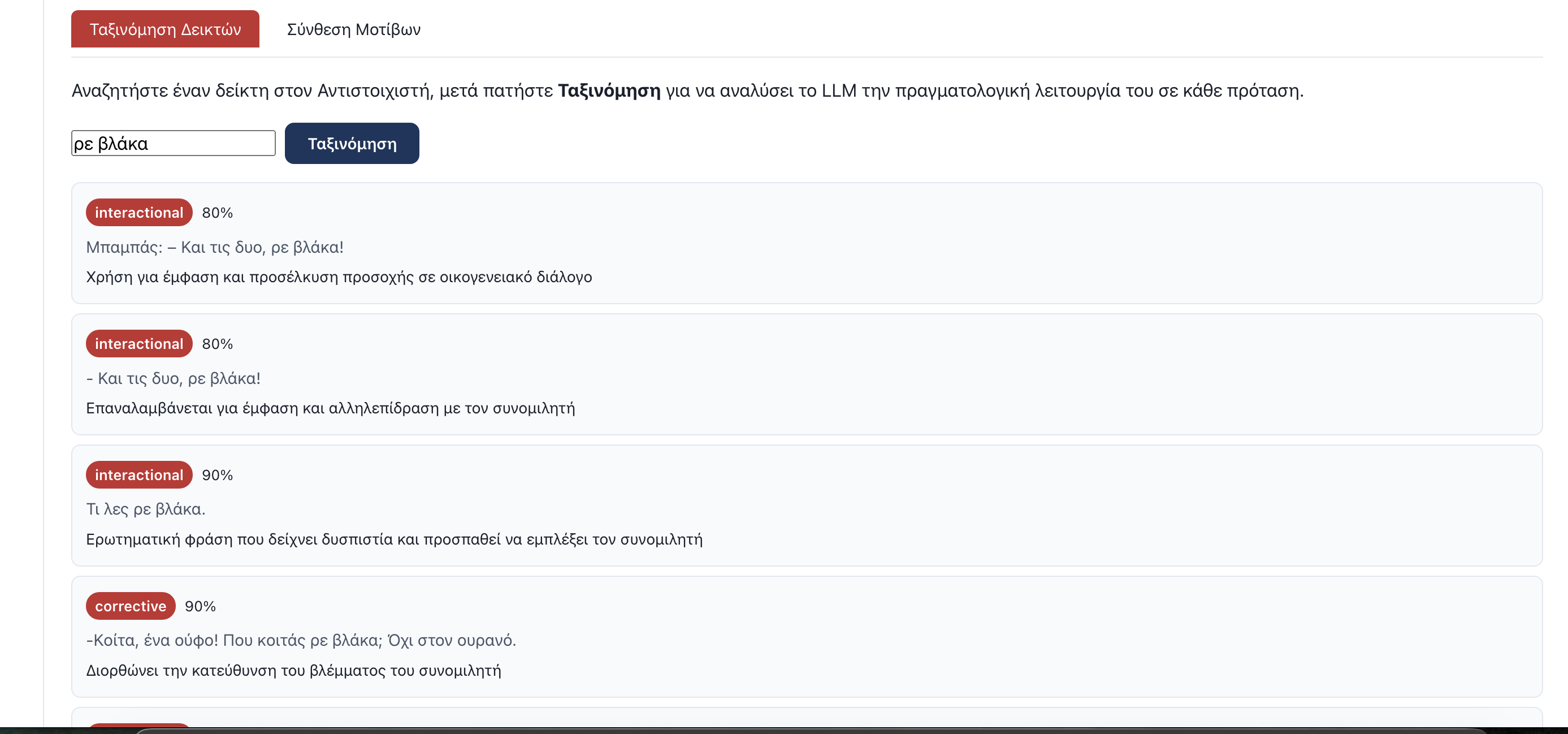}
		
		\caption{LLM layer applied to actual classification. For context-extracted markers (here, the interaction marker ρε), the model classifies the actual features of each occurrence into confidence scores. These confidence scores are based on actual match rows, not the model's parameter memory.}
		
		\label{fig:llm}
		
	\end{figure}
	\section{Usage Examples}\label{sec:examples}
	
	To illustrate the range of queries Svarna supports, we present three brief examples.
	
	\subsection{Cross-register Distribution of a Discourse Marker}
	
\begin{figure}[htbp]
		\centering
		\includegraphics[width=0.85\textwidth]{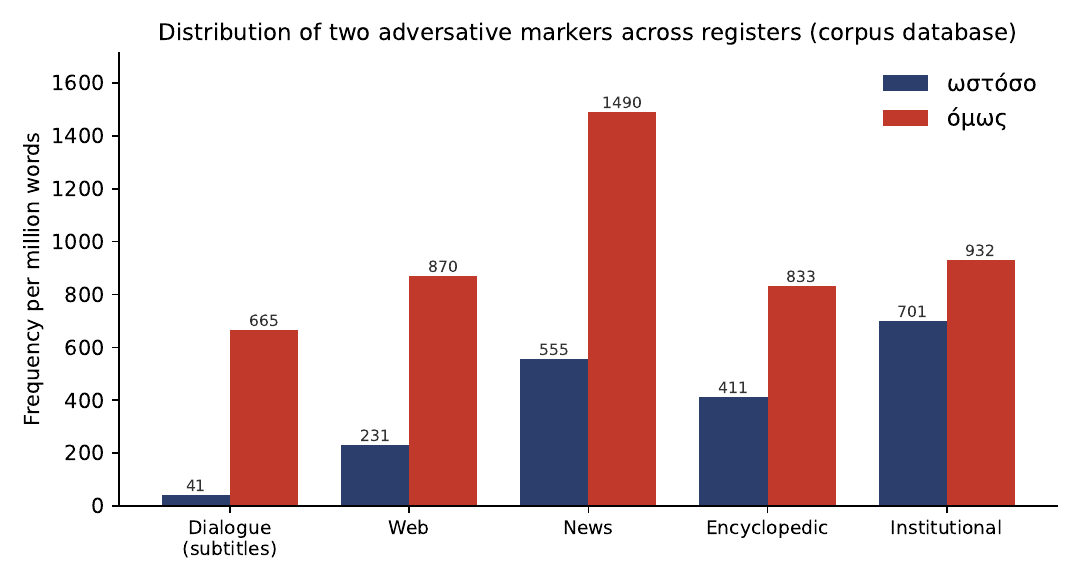}
		\caption{Frequency per million words of the adversative markers ωστόσο and όμως across registers in the corpus database. ωστόσο is concentrated in formal registers (institutional, news, encyclopedic) and almost absent from subtitle dialogue, whereas όμως is frequent across all registers, consistent with ωστόσο being a formal-register marker and όμως a register-neutral one.}
		\label{fig:markers}
	\end{figure}

	Consider the adversative marker ωστόσο (`however'). Using the discourse markers tab, we can immediately see that this marker occurs with high frequency in Wikipedia and parliamentary text but is virtually absent from subtitles and dialectal data. This distributional pattern is consistent with ωστόσο being a formal register marker, as opposed to the more colloquial όμως, which shows a flatter distribution across registers. Such observations, which in principle require hours of manual corpus work, are available instantly through the pre-computed marker frequency table (Figure~\ref{fig:markers}).
	
	\subsection{Dialectal Comparison}
	
	Searching for the copula form είναι in the dialectal database reveals variation in frequency across dialect groups. Cretan and Cypriot texts show high frequencies, consistent with productive use of the standard copula. Some dialect groups show lower relative frequency, potentially reflecting the use of dialectal variants. The KWIC display allows the researcher to inspect individual examples in context and the metadata panel identifies the specific text or speaker for each hit.
	
	\subsection{Collocation Analysis of a Polysemous Word}
	
	The word βαρύ (`heavy') shows different collocational profiles depending on the register. In parliamentary text, its top collocates include φορολογικό (`tax') and κόστος (`cost'), reflecting metaphorical usage. In literary text, collocates shift toward physical and emotional senses. The MI scores \citep{church1990} help distinguish genuinely associated collocates from high-frequency co-occurrences, and the collocation network visualization provides an overview of the word's associative field.

	\section{Discussion}\label{sec:discussion}
	
	Svarna is closer to a practical tool than a theoretical contribution. Its value lies in the fact that it brings existing materials together and allows them to be searched through a single interface. However, several limitations must also be considered.
	
	First, the materials included in Svarna were collected in an opportunistic manner. Because it includes materials that are freely available and redistributable, it cannot be considered a balanced collection in the traditional sense. Wikipedia and parliamentary minutes are included excessively compared to everyday conversations or personal correspondence. While the dialect database includes nine dialect groups, this reflects the availability of transcribed dialect data rather than a principled sampling strategy.
	
	Second, this system performs all analysis at the time of query based on matching samples rather than the entire corpus. In other words, the frequency of very common words is an approximation, and collocation statistics are based on samples rather than the entire corpus. While this is sufficient for most practical purposes, it may fall short compared to systems like Sketch Engine, which pre-calculates word sketches based on fully parsed data.
	
	Finally, in the current version, adding a corpus requires self-hosting. This means you must clone the repository, run build scripts on your own data, and deploy a local instance. This is possible thanks to the open architecture and generous licensing, but it requires a certain level of technical capability. The next step is to build an in-app upload interface. This allows users to directly add text and create indexes via the web frontend without separate local configuration.
	
	Despite these limitations, Svarna fills a significant gap. To our knowledge, Svarna is the only free, open-source, web-based tool capable of searching over 507 million words of Greek text across various contexts (institutions, literature, dialects, social media, historical records, etc.) and performing various tasks useful for corpus linguistics.

	\section{Conclusion}\label{sec:conclusion}
	
	This paper introduced Svarna, an open corpus workbench for Modern Greek. Svarna integrates institutional, literary, dialect, social media, and historical corpora into a single searchable platform. This system provides concordancing, frequency analysis, collocation extraction, discourse marker profiling, text analysis, register comparison, and regular expression search capabilities, all of which can be accessed via a web browser without installation or credentials.
	
	Although we implemented Svarna for Modern Greek in this article, no part of its architecture is specific to any language. The FTS5 \texttt{unicode61} tokenizer, single table schema, and build pipeline are language-independent. Since the same workbench can be reused for any language simply by specifying a new text set in the build script, Svarna can be utilized as a general and reusable template for low-cost corpus platforms.
	
	The name Svarna signifies a traditional leveling tool. This study is the first to survey available materials related to Modern Greek. The materials are uneven, the scope is incomplete, and there is still much work to be done. However, since this field is open and the tools are open, we look forward to better sweeps in the future.

\section*{Acknowledgements}

The author thanks the developers and maintainers of the open data resources that made this work possible, including Project Gutenberg, the Greek Parliament Proceedings Dataset, the GRDD+ project, the BabyLM shared task, and the Universal Dependencies initiative. The author also wishes to thank Microsoft Azure for providing computational support for parts of this project through a grant to support the project ``A Computational Dialectal Atlas of Greek: LLMs and Educational Applications for Greek Dialects'' (CoDAG).

\section*{Declarations}

\begin{description}
	\item[Funding] Microsoft Lingua grant scheme:  ``A Computational Dialectal Atlas of Greek: LLMs and Educational Applications for Greek Dialects'' (CoDAG)
	\item[Competing interests] The author declares no competing interests.
	\item[Data availability] The source corpora are derived from publicly available datasets cited in the paper. The database build scripts are included in the source code repository. Pre-built databases will be made available through a data repository upon publication.
	\item[Code availability] The source code, build scripts, and deployment configuration are available under the MIT license at \url{https://github.com/StergiosCha/SVARNA_CORPUS_APP}.
	\item[Author contribution] S. Chatzikyriakidis designed the system, built the corpora, implemented the software, deployed the platform, and wrote the paper.
\end{description}

\bibliographystyle{apalike}
\bibliography{references}

\end{document}